# Crowd Management in Open Spaces

Tauseef Ali, University of Twente, The Netherlands
Ahmed B. Altamimi

*Abstract*—Crowd analysis and management is a challenging problem to ensure public safety and security. For this purpose, many techniques have been proposed to cope with various problems. However, the generalization capabilities of these techniques is limited due to ignoring the fact that the density of crowd changes from low to extreme high depending on the scene under observation. We propose robust feature based approach to deal with the problem of crowd management for people safety and security. We have evaluated our method using a benchmark dataset and have presented details analysis.

## I. INTRODUCTION

Machine learning and computer vision methods [1][2][3] present different techniques how to handle crowded scenes, in real time, in different public places including but not limited to museums, concerts, airports, and stadiums. The development of these methods assist safety and security personnel to take important measures to elevate efficiency and profitability and giving their customers better experience at the different public spaces. The aforementioned methods [4][5] execute video frames input from commercial off the shelf CCTV cameras.

In the public scenes [6][7], the individual tracking method [25] considers people counting, queue analysis, congestion detection, stampede identification, and people flow analysis by exploiting advance deep learning and machine learning approaches. Considering these advances, a hybrid approach, according to the methods [8][9][10], can be developed that renders live people analytics about everything from flows, queues and wait times to processing times, occupancies, and asset utilization considering different people gathering spaces. This kind of hybrid method could be exploited to model measurable insights to enhance and assist in real-time operations at public places. The data collected by the authors from public spaces [11][12][13] present important dossier for future planning and investment decisions.

## II. RELATED WORK

Considering different public places, a tracking based method [14][15][33] can be designed for customer and individual tracking and queue analysis to enhance the capability of checkpoint efficiency. This kind of method would assist to reduce queues and waiting times. Additionally, these methods also assist public spaces management personnel to take care of customers into retail areas more swiftly. Moreover, it can also be exploited to improve customer tracking capabilities to understand how passengers move around retail and food/beverage areas [16][17][26]. By considering and evaluating dwell times and retail conversion metrics, the management can enhance and increase the size of their retail zones.

Apart from those methods, we can also model granular, real-time customer tracking analysis regarding flows [18][19], queue lengths, public count, and wait times, processing times, and resources utilization. To considering these problems, both the customers/public and staff can be interviewed. Evaluating people tracking information is the first step in enhancing staffing at checkpoints, enhancing the quality of queue management and efficiency, and pushing passengers swiftly and happily into revenue oriented areas. This feedback also provides opportunities for the public spaces management personnel to establish stronger and more productive relationships with its related companies and industrial partners. Additionally, computer vision methods [20][21][22] assist public spaces personnel with improved levels of situational awareness, and the chance to cope with different problems by exception, as they show up. For example, real time high crowded level detection system can be used to deal with safety and security issues more quickly [23][24].

Ullah and Alaya [25] proposed a vision based graphical model for multi-target tracking in very challenging environments. Their method performs exhaustive search to find a set of robust tracks for different moving targets explicitly not mentioning any heuristics or relaxation. Ullah and Saqib et al. [3][7][4][27] investigated methods for moving objects segmentation and anomalous situation detection in public spaces. Their methods present effective abilities to deal with different public places having various properties including diverse densities, sparseness, and coherence. For the same purpose, Bisagno et al. [8] consider the layout of the physical environments, and the availability of mutual occlusions and obstacles. Older prediction approaches use anticipating individual's future path based on the precise motion history of an object. They consider that conventional tracking methods are not good enough to track in populated and crowded situations. Additionally, those approaches are not working in real-time to be consider for real environments, immediately. From social perspective, it is common observation that people including friends, couples, family members present coherent motion patterns when they are walking in groups. Considering this reality, the researchers propose an effective technique to search future trajectories in populated environments, at the group level. In the beginning, they explore the consistency in motion as a whole, and combine trajectories that are characterized by similar properties. The method help to segment people within the same social group. Hao et al. [17] present two significant advancements consisting of a spatial-temporal texture

Tauseef Ali is with the University of Twente, the Netherlands. He is working with the datamanagement and biometrics group.

extraction approach which is able to effectively get video and image textures with rich movement patterns. They exploit Gabor filter textures with the strongest entropy values. They also use a stronger method for detecting people motion patterns and to design and identify anomalous behaviors in the public spaces by using an enhanced gray level co-occurrence matrix model. Kang et al. [21] proposed that the accuracy of individual-based methods reduces when the resolution of image and video frames are low quality. The same accuracy decreases when there are stronger occlusions. Taking into account these counting methods, they ignore individuals localization and adopt regression-based approaches to directly count people in dense public places. Considering regression-based methods, density map calculations, where the count of people within an region of interest (ROI) is the sum of the density map over that ROI, is presenting improved results because it considers spatial properties, because of which it is robust for both public counting and detection. Due to the significant performance improvement of deep convolutional neural networks, the counting performance has gravitated to greater extent. Kang et al. evaluated density maps engendered by density estimation methods considering different pedestrian analysis issues including but not limited to counting, localization, and tracking. The famous CNN methods produce density maps with resolution that is less than the original video frames, because of the downsample strides in the convolution/pooling steps. To provide proper density map, they also evaluate a traditional CNN that uses a sliding window regressor to search the density for every pixel in the image or a frame of a video. They use a fully convolutional adaptation, with skip connections from initial layers to consider loss in spatial properties during upsampling.

Shimura et al. [19] investigate that in some social or public events, people make around about motion circulating around the central part of the environment or a physical obstacle. They propose a popular method to copy such environment. Additionally, these researchers use cellular automata where coding is performed in square lattice with static floor fields in polar coordinate. To copy and model the same social behavioral representation, the researchers use the floor field associated with the wall avoidance and lane formation models. The outcomes of the results are evaluated to analyze the circularity of the people movement by considering the optical flow technique, one of the popular technique in the field of computer vision and machine learning. They considered their method using velocity vector of the brightness properties to evaluate the simulated videos. The results of different simulations outcomes show that this method accurately regenerate the roundabout movement of people in various crowd scenes.

III. PROPOSED METHOD

For crowd management and event of interest detection, we consider motion atoms [28] form the scene. Motion atoms demonstrate complex and unique motion patterns, and can be observed as mid-level characteristics to combine the gap between low-level features and high-level features in order to encode complex events. A single motion atom presents motion information in a short temporal window. The unique ability of motion atom is characterized by the size of the temporal window. We observed that the temporal layout (i.e. sequential structure of motion atoms), encodes motion information in a longer scale and presents significant hint to identify different events in the crowds. It is worth noticing that the number of candidate fusion is increasing significantly with the number of atomic representations and much of the fusions are not very coherent for classification.

Considering these observations, we use motion atom and phrase inspired by the work of [28] which is associated with a mid-level representation of crowded scenes, which jointly captures the motion, appearance, and temporal layout of complex crowd events. According to the method, we investigate a set of motion atoms from crowd videos consisting of complex events. The crowd videos only consist of class labels. We considering it as a unique clustering challenge. An iterative algorithm is modeled, which switches between clustering segments and training classifier for each cluster. In this method, each cluster represents a motion atom. In fact, the atoms in our method are the building blocks for event representation in the crowd videos. Moreover, we design motion phrase as a temporal fusion of many motion atoms. This process encodes both short-scale motion data of individual atom and encodes the temporal layout of many atoms in a longer temporal window. For this purpose, we consider an AND/OR paradigm to describe motion phrase, allowing us to cope with temporal displacement efficiently. In fact, the method introduces a bottom-up technique and greedy approach to extract motion phrases with high uniqueness and discriminative ability. The approach describe each video by the algebraic vector of motion atoms and phrases by max pooling the response matrix of each atom and phrase in the crowd videos.

The provided input to out method is a refine set of crowd events videos. It is important to note that we overlook the class label of each event in the crowd scene. We heavily extract segments from each crowd video consisting of complex events. Because of the nature of the crowd video that each training crowd video is a small video clip that is randomly consistent with temporal window, we classify each short clip into k segments of equal window with the possibility of 50% overlaps. For further motion analysis, we obtain dense trajectory features according to the technique outline in [29] for each crowd video segment. We then consider the Bag of Visual Words technique [30] to describe each crowd video segment. For this purpose, we select four types of features including HOG, HOF, MBHX, and MBHY. For each feature, we model a codebook of size K, and a histogram representation is represented for each segment. For the purpose of designing and making group segments, we entail to describe a similarity model between segments. Given two segments $S_i$ and $S_j$, we formulate their similarity according to the equation

$$\text{Sim}(S_i, S_j) = \sum_{m=1}^{4} \exp(-\mathcal{D}(\mathbf{h}_i^m, \mathbf{h}_j^m)),$$

Where D(h, h) is the normalized distance between the two histograms according to the equation

$$\mathcal{D}(\mathbf{h}_i^m, \mathbf{h}_j^m) = \frac{1}{2M_m} \sum_{k=1}^{D} \frac{(\mathbf{h}_{i,k}^m - \mathbf{h}_{j,k}^m)^2}{\mathbf{h}_{i,k}^m + \mathbf{h}_{j,k}^m},$$

Where $h_i^m$ represents the histogram vector for feature channel of crowd video segment $S_i$. Similarly, $M^m$ is the average distance for channel $m$ over the crowd events training samples align with their corresponding labels. Moreover, a motion phrase P is unique for c-th class of a crowd event if it is very consistent with this class, but shows untouched in other crowd event classes. We describe its uniqueness capability as formulated in the equation.

$$\text{Dis}(P, c) = \text{Rep}(P, c) - \max_{c_i \in C-c} \text{Rep}(P, c_i),$$

In the equation, C shows all the classes and Rep(P; c) represents the strong capability of P in accordance with class c, whose greater magnitude shows powerful coherency with the class c. Additionally, in the equation

$$\text{Rep}(P, c) = \frac{\sum_{i \in S(P,c)} r(V_i, P)}{|S(P, c)|},$$

The term (Vi; P) shows the response magnitude of motion phrase P in video Vi and S(P; c) is a bunch of crowd videos formulated in the equation.

$$S(P, c) = \{i | \text{Class}(V_i) = c \land V_i \in \text{top}(P)\},$$

In the above equation, Vi represents the class label of video Vi and top(P) shows a bunch of crowd clips that have the greater return magnitudes for motion phrase P. Because of the huge dissimilarities among different crowd events, a single motion phrase could get greater magnitude only on elements of the clips of certain event.

For the classification and recognition of events, we use support vector machine (SVM) [31][34]. If we are provided training data which might be, for example, in our case different events of interest in crowded scenes. Our objective is to identify a function that represents at most ε deviation from the actually chosen targets for all the training events, and at the same time is as flat as possible to be best trained according to the available data. To say the same in different statements, we do not take into account errors as long as they are less than ε, but we will not embrace any deviation larger than this the presented error. This is significant since if we want to be sure not to lose more than ε classification error when coping with different events in the crowded scenes. For simplicity, we start by mentioning the case of linear functions f shaping it according to the equation.

$$f(x) = \langle w, x \rangle + b \text{ with } w \in \mathcal{X}, b \in \mathbb{R}$$

where $\langle \cdot, \cdot \rangle$ shows the dot product. Flatness in the above equation means that one finds a small w. The right technique is to ensure this is to minimize the norm. This could be treated as convex optimization issue formulated according to the equation.

$$\begin{aligned} \text{minimize} \quad & \tfrac{1}{2}\|w\|^2 \\ \text{subject to} \quad & \begin{cases} y_i - \langle w, x_i \rangle - b \leq \varepsilon \\ \langle w, x_i \rangle + b - y_i \leq \varepsilon \end{cases} \end{aligned}$$

The strongest assumption in the equation is that such a function rarely available that computes all pairs of training samples with their corresponding labels and with ε accuracy, or in other terminology, that the convex optimization issue is possible.

$$\begin{aligned} \text{minimize} \quad & \frac{1}{2}\|w\|^2 + C \sum_{i=1}^{\ell}(\xi_i + \xi_i^*) \\ \text{subject to} \quad & \begin{cases} y_i - \langle w, x_i \rangle - b \leq \varepsilon + \xi_i \\ \langle w, x_i \rangle + b - y_i \leq \varepsilon + \xi_i^* \\ \xi_i, \xi_i^* \geq 0 \end{cases} \end{aligned}$$

The constant C > 0 finds the trade-off between the flatness off and the extent to which deviations larger than ε are acceptable.

## IV. EXPERIMENTS

We assess our method on UCSD dataset [32], which is a properly annotated benchmark for the analysis of abnormal detection and localization in crowded scenes. The dataset was captured with a CCTV camera fixed at an elevation at a resolution of 238×158 with 10 fps, overlooking people pathways. The movement of non-people objects in the pathways, and anomalous pedestrian motion patterns are treated as abnormal situations. In general, these anomalies are bikers, skaters, small carts, and people walking across a pathway or in the park. Videos clips in the dataset are divided into 2 subsets: Ped1 and Ped2, each one is associated with a different scenario. These videos recorded from each scene were categorized into different clips each of which has around 200 frames. Ped1 consists of 34 training videos and 36 testing videos. Ped2 consists of 16 training videos and 14 testing videos. For each video, the ground truth annotation includes a binary flag per frame, showing whether an anomalous entity is present in that particular video frame. Moreover, there is a subset consisting of 10 videos is available with manually produced pixel-level binary masks, which recognize the parts or sub-parts consisting of abnormal events with clear boundaries. This is generated for the analysis of performance with respect to the ability of anomaly localization and segmentation. Sample image from UCSD dataset is shown.

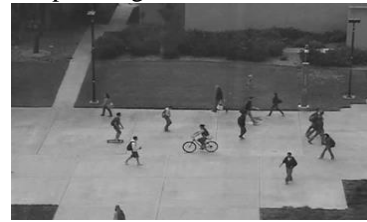

For experimental analysis, Table 1 presents the area under ROC curve (AUC) of the tested methods, in which a larger AUC score shows improved classification results. Table1 shows that our method outputs competitive results against many popular reference methods.

TABLE I: RESULTS OF REFERENCE AND OUR METHODS

| Methods | MDT | SF | MPPCA | Ours |
|---|---|---|---|---|
| AUC | 0.78 | 0.74 | 0.65 | 0.79 |

V. CONCLUSION

In this paper, we presented a novel method for detecting interesting and anomalous events in crowded scenes. We proposed a holistic method that is independent of crowd density. It does not matter how concentrated crowd is in certain locations of the scene. We tested our method on UCSD dataset which is a benchmark dataset.

REFERENCES

[1] Khan, S. D., Tayyab, M., Amin, M. K., Nour, A., Basalamah, A., Basalamah, S., & Khan, S. A. (2017). Towards a Crowd Analytic Framework For Crowd Management in Majid-al-Haram. arXiv preprint arXiv:1709.05952.

[2] Ahmad, K., Conci, N., & De Natale, F. G. (2018). A saliency-based approach to event recognition. Signal Processing: Image Communication, 60, 42-51.

[3] Ullah, H., Altamimi, A. B., Uzair, M., & Ullah, M. (2018). Anomalous entities detection and localization in pedestrian flows. Neurocomputing, 290, 74-86.

[4] Saqib, M., Khan, S. D., Sharma, N., & Blumenstein, M. (2017, December). Extracting descriptive motion information from crowd scenes. In 2017 International Conference on Image and Vision Computing New Zealand (IVCNZ) (pp. 1-6). IEEE.

[5] Khan, W., & Ullah, H. (2010). Authentication and Secure Communication in GSM, GPRS, and UMTS Using Asymmetric Cryptography. International Journal of Computer Science Issues (IJCSI), 7(3), 10.

[6] Saqib, M., Khan, S. D., & Blumenstein, M. (2016, November). Texture-based feature mining for crowd density estimation: A study. In Image and Vision Computing New Zealand (IVCNZ), 2016 International Conference on (pp. 1-6). IEEE.

[7] Ullah, H., Ullah, M., & Uzair, M. (2018). A hybrid social influence model for pedestrian motion segmentation. Neural Computing and Applications, 1-17.

[8] Bisagno, N., Zhang, B., & Conci, N. (2018, September). Group LSTM: Group Trajectory Prediction in Crowded Scenarios. In European Conference on Computer Vision (pp. 213-225). Springer, Cham.

[9] Ahmad, F., Khan, A., Islam, I. U., Uzair, M., & Ullah, H. (2017). Illumination normalization using independent component analysis and filtering. The Imaging Science Journal, 65(5), 308-313.

[10] Ullah, H., Uzair, M., Ullah, M., Khan, A., Ahmad, A., & Khan, W. (2017). Density independent hydrodynamics model for crowd coherency detection. Neurocomputing, 242, 28-39.

[11] Trabelsi, R., Jabri, I., Melgani, F., Smach, F., Conci, N., & Bouallegue, A. (2017, November). Complex-Valued Representation for RGB-D Object Recognition. In Pacific-Rim Symposium on Image and Video Technology (pp. 17-27). Springer, Cham.

[12] Ullah, M., Ullah, H., & Alseadonn, I. M. (2017). Human action recognition in videos using stable features.

[13] Xu, M., Ge, Z., Jiang, X., Cui, G., Zhou, B., & Xu, C. (2019). Depth Information Guided Crowd Counting for Complex Crowd Scenes. Pattern Recognition Letters.

[14] Alameda-Pineda, X., Ricci, E., & Sebe, N. (2019). Multimodal behavior analysis in the wild: An introduction. In Multimodal Behavior Analysis in the Wild (pp. 1-8). Academic Press.

[15] Ullah, M., Ullah, H., Conci, N., & De Natale, F. G. (2016, September). Crowd behavior identification. In Image Processing (ICIP), 2016 IEEE International Conference on (pp. 1195-1199). IEEE.

[16] Kim, H., Han, J., & Han, S. (2019). Analysis of evacuation simulation considering crowd density and the effect of a fallen person. Journal of Ambient Intelligence and Humanized Computing, 1-11.

[17] Hao, Y., Xu, Z. J., Liu, Y., Wang, J., & Fan, J. L. (2019). Effective crowd anomaly detection through spatio-temporal texture analysis. International Journal of Automation and Computing, 16(1), 27-39.

[18] Ullah, H., Ullah, M., Afridi, H., Conci, N., & De Natale, F. G. (2015, September). Traffic accident detection through a hydrodynamic lens. In Image Processing (ICIP), 2015 IEEE International Conference on (pp. 2470-2474). IEEE.

[19] Shimura, K., Khan, S. D., Bandini, S., & Nishinari, K. (2016). Simulation and Evaluation of Spiral Movement of Pedestrians: Towards the Tawaf Simulator. Journal of Cellular Automata, 11(4).

[20] Ullah, H. (2015). Crowd Motion Analysis: Segmentation, Anomaly Detection, and Behavior Classification (Doctoral dissertation, University of Trento).

[21] Kang, D., Ma, Z., & Chan, A. B. (2018). Beyond counting: Comparisons of density maps for crowd analysis tasks-counting, detection, and tracking. IEEE Transactions on Circuits and Systems for Video Technology.

[22] Rota, P., Ullah, H., Conci, N., Sebe, N., & De Natale, F. G. (2013, September). Particles cross-influence for entity grouping. In Signal Processing Conference (EUSIPCO), 2013 Proceedings of the 21st European (pp. 1-5). IEEE.

[23] Ullah, H., Ullah, M., & Conci, N. (2014, March). Real-time anomaly detection in dense crowded scenes. In Video Surveillance and Transportation Imaging Applications 2014 (Vol. 9026, p. 902608). International Society for Optics and Photonics.

[24] Arif, M., Daud, S., & Basalamah, S. (2013). Counting of people in the extremely dense crowd using genetic algorithm and blobs counting. IAES International Journal of Artificial Intelligence, 2(2), 51.

[25] Ullah, M., & Alaya Cheikh, F. (2018). A Directed Sparse Graphical Model for Multi-Target Tracking. In Proceedings of the IEEE Conference on Computer Vision and Pattern Recognition Workshops (pp. 1816-1823).

[26] Khan, W., Ullah, H., & Hussain, R. (2013). Energy-Efficient Mutual Authentication Protocol for Handhled Devices Based on Public Key Cryptography. International Journal of Computer Theory and Engineering, 5(5), 754.

[27] Ullah, M., Mohammed, A., & Alaya Cheikh, F. (2018). PedNet: A Spatio-Temporal Deep Convolutional Neural Network for Pedestrian Segmentation. Journal of Imaging, 4(9), 107.

[28] Wang, LiMin, Yu Qiao, and Xiaoou Tang. "Mining motion atoms and phrases for complex action recognition." Proceedings of the IEEE international conference on computer vision. 2013.

[29] Wang, Heng, Alexander Kläser, Cordelia Schmid, and Cheng-Lin Liu. "Dense trajectories and motion boundary descriptors for action recognition." International journal of computer vision 103, no. 1 (2013): 60-79.

[30] Wang, Xingxing, LiMin Wang, and Yu Qiao. "A comparative study of encoding, pooling and normalization methods for action recognition." In Asian Conference on Computer Vision, pp. 572-585. Springer, Berlin, Heidelberg, 2012.

[31] Smola, Alex J., and Bernhard Schölkopf. "A tutorial on support vector regression." Statistics and computing 14, no. 3 (2004): 199-222.

[32] Mahadevan V, Li W, Bhalodia V, Vasconcelos N (2010) Anomaly detection in crowded scenes. In: IEEEconference on computer vision and pattern recognition (CVPR), pp 1–8.

[33] Khan, Sultan Daud, and Habib Ullah. "A survey of advances in vision-based vehicle re-identification." Computer Vision and Image Understanding (2019).

[34] Ullah, Habib, Muhammad Uzair, Arif Mahmood, Mohib Ullah, Sultan Daud Khan, and Faouzi Alaya Cheikh. "Internal Emotion Classification Using EEG Signal with Sparse Discriminative Ensemble." IEEE Access (2019).